# The negation of permutation mass function

Yongchuan Tang [1,2,*], Rongfei Li [1]

1. School of Microelectronics, Northwestern Polytechnical University, Xi'an, Shaanxi 710072, China

2. Chongqing Innovation Center, Northwestern Polytechnical University, Chongqing, 401135, China

∗ Corresponding author at: School of Microelectronics, Northwestern Polytechnical University, Xi'an, Shaanxi 710072, China.

E-mail address: tangyongchuan@nwpu.edu.cn (Y. Tang).

## Abstract

Negation is an important perspective of knowledge representation. Existing negation methods are mainly applied in probability theory, evidence theory and complex evidence theory. As a generalization of evidence theory, random permutation sets theory may represent information more precisely. However, how to apply the concept of negation to random permutation sets theory has not been studied. In this paper, the negation of permutation mass function is proposed. Moreover, in the negation process, the convergence of proposed negation method is verified. The trends of uncertainty and dissimilarity after each negation operation are investigated. Numerical examples are used to demonstrate the rationality of the proposed method.

## 1 Introduction

A profusion of information saturates our daily lives. With the advancements in sensors and artificial intelligence, the processing and analysis of information have become particularly crucial. In the real world, information typically manifests with inherent uncertainties[1][2]. Researchers have proffered numerous theories aimed at the processing and modeling of uncertain information. Such as intuitionistic fuzzy set theory[3][4], Z-number theory[5][6], evidential reasoning[7][8], probability theory (PT)[9], evidence theory (ET)[10][11], random permutation sets theory (RPS)[12][13] and so on[14][15]. These theories are widely applied in information fusion[16][17], pattern classification[18][19], complex network analysis[20][21] and so on[22][23].

Among these theories, the essence of probability theory, evidence theory, and random permutation set theory is similar, as they all involve the allocation of belief within specific event spaces. Probability theory, evidence theory, and random permutation sets represent belief assignments by probability distribution (PD), basic probability assignment(BPA), and permutation mass function (PM), respectively. In probability theory, credibility is allocated to independent mutually exclusive events, while evidence theory expands the credibility allocation space to encompass both independent mutually exclusive events and their combinations. random permutation sets were initially proposed by Deng et al[13], representing a further extension of evidence theory. It simultaneously considers the random combinations and permutation orders of events, enabling a more refined modeling of uncertain information. Moreover, evidence theory and random permutation sets theory can be compatible with probability theory under certain conditions. Due to its excellent knowledge modeling capabilities, random permutation sets have gradually became as a hot topic in contemporary research.

Besides, the representation and processing of information remain an open issue. "Negation" is an important method of information representation, as it provides a perspective akin to the "opposite" of information. For instance, demonstrating the validity of a proposition can be challenging, whereas proving its falsity requires only the presentation of a single counterexample. This concept is exemplified in mathematics by the principle of "proof by contradiction." Additionally, the uncertainty of a particular piece of information can also be assessed by measuring the discrepancy (conflict) between the information itself and its negation. The greater the discrepancy between the original information and its negation, the lower the uncertainty (fuzziness) associated with that information. The process of "negation" can be seen as a bridge from the positive aspect of an event to its negative aspect. Zadeh formally introduced the concept of "negation" in probability theory on his BISC blog, sparking widespread interest among researchers. Smets employed matrix methods to investigate how to determine the negation of belief functions[24]. Yager introduced an approach to obtain the negation of probability distribution with maximal entropy[25]. Zhang et al extended the Yager's negation method from the aspect of Tsallis entropy[26]. Since evidece theory can be considered as a generalization of probability, inspired by Yager's work, Yin et al applied the concept of "negation" to basic probability assignment(BPA) in evidence theory[27]. Luo et al introduced a novel definition of negation BPA by using matrix operator[28]. Under the quantum model of evidence theory, Xiao et al studied the negation of quantum mass function[29]. Liu et al proposed the negation of discrete Z-numbers based on the combination of probability and fuzziness[30].

As previously mentioned, similar to probability theory and evidence theory, random permutation sets theory is a method for uncertain modeling. However, the negation of permutation mass function has not been studied. To address this issue, the definition of "negation permutation mass function" is proposed in this paper, which is a new approach to represent and process the knowledge based on random permutation sets theory. In addition, based on Chen[31] and Deng's [32]work, the change of entropy and dissimilarity of permutation mass function after each iteration of the negation process is presented.

The rest of this paper is organized as follows. In Section 2, the related preliminaries is briefly presented. In section 3, the negation of *PM* is proposed. The convergence, entropy and dissimilarity of *PM* during the negation process are analyzed in Section 4. Finally, Section 5 concludes this work.

## 2 Preliminaries

### 2.1 Dempster-Shafer evidence theory
#### 2.1.1 Frame of discernment(FOD)

Frame of discernment is a set contains all the observable events. Assuming there are $n$ mutually exclusive events, the corresponding FOD can be represent as $\Phi = \{\varphi_1, \varphi_2, \cdots \varphi_n\}$. The power-set of $\Phi$ is defined as

$$2^\Phi = \{\varnothing, \{\varphi_1\}, \cdots, \{\varphi_n\}, \{\varphi_1, \varphi_2\}, \cdots, \Phi\}$$

#### 2.2.2 Basic probability assignment(BPA)/mass function

The basic probability assignment(BPA)/mass function is a mapping:

$$m: 2^\Phi \to [0,1]$$

where $m(\varnothing) = 0$, $\sum_{A \in 2^\Phi} m(A) = 1$.

If $m(A) > 0$, $A$ is called focal element.

## 2.2 Random permutation theory
### 2.2.1 Permutation event space(PES)

Similar to the evidence theory, supposing set $\Theta$ contains $n$ mutually exclusive events $\Theta = \{\gamma_1, \gamma_2, \cdots \gamma_n\}$. The corresponding permutation event space contains all the possible permutations of the elements in $\Theta$, denoted as

$$PES(\Theta) = \{M_{rs} | r = 0,1,2,\cdots,n; s = 1,2,\cdots,P(n,r)\}$$
$$= \{\varnothing, \{\gamma_1\},\cdots,\{\gamma_{n-1}\},\{\gamma_n\},\{\gamma_1,\gamma_2\},\{\gamma_2,\gamma_1\},\cdots,\{\gamma_{n-1},\gamma_n\},\{\gamma_n,\gamma_{n-1}\},\cdots,$$
$$\{\gamma_1,\gamma_2,\cdots,\gamma_n\},\cdots\{\gamma_n,\gamma_{n-1},\cdots\gamma_1\}\}$$

where, $r$ is the cardinality of subset $M_{rs}$, $s$ is the the existence of $s$ distinct r-permutations, $P(n,r)$ represents the $r$ permutation of $n$, it can be calculated as $P(n,r) = \frac{n!}{(n-r)!}$, the cardinality of $PES(\Theta)$ is denoted as $\Delta = \sum_{i=0}^{N} P(n,r)$.

### 2.2.2 Permutation mass function/random permutation sets theory/Permutation focal element

A permutation mass function is a mapping: $PM: PES(\Theta) \to [0,1]$, where $PM(\varnothing) = 0$, $\sum_{A \in PES(\Theta)} PM(A) = 1$.

If $PM(A) > 0$, $A$ is called permutation focal element.

The random permutation sets(RPS) is a set of pairs, which is defined as:
$$RPS(\Theta) = \{\langle A, PM(A)\rangle | A \in PES(\Theta)\}$$

*PM(A)* in the random permutation sets can be regarded as the belief assignment on *A*, which is similar to the meaning of mass function in evidence theory. Because $PES(\Theta)$ not only includes random combinations between events $\gamma_i$, but also considers their permutation order, the belief in random permutation sets theory is assigned to a more refined space than probability theory and evidence theory, which has stronger information representation ability. Consider an target recognition scenario. There are 3 targets *A*, *B* and *C*. In probability theory, probability theory can only express the probability of any one of *A*, *B* or *C*, but can not express the situation that there are multiple targets at the same time. The evidence theory treats the multi-target situation as an independent situation, instead of adding the single-target's belief directly(i.e., $m(A,B)$ may not equals to $m(A) + m(B)$), which improves this problem to some extent. On the basis of evidence theory, the random permutation sets also considers the order of the appearance of the target, and assigns the belief to the different order of the appearance of the target independently(i.e., $PM(A,B,C)$ may not equals to $PM(C,B,A)$, because event (A, B, C) is different from (C, A, B) ).

Figure 1 illustrates the differences in the belief assignment space(event space) for different theories in this target recognition example. It should be highlighted that the random permutation sets theory can degenerate into evidence theory and probability theory under certain circumstances.

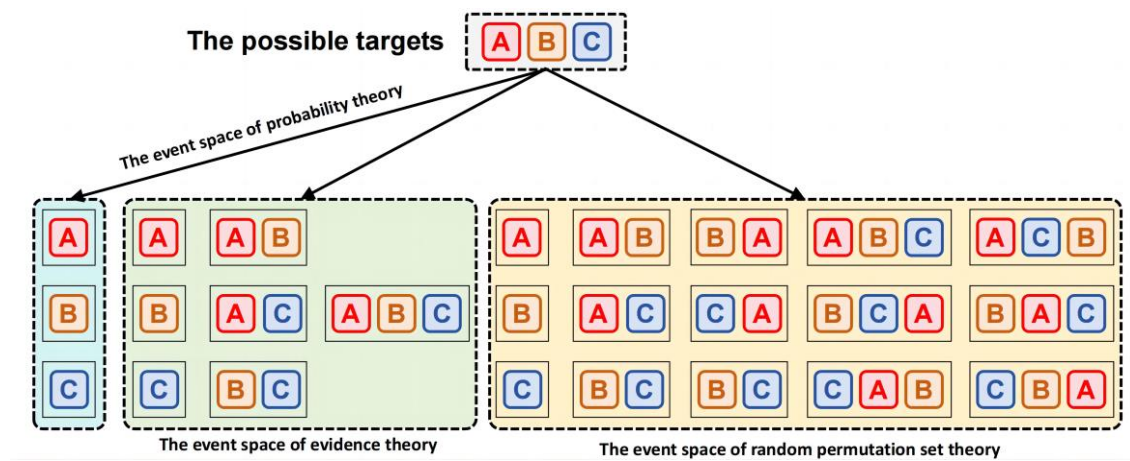

Fig. 1 The illustration of belief assignment space(event space) of different theories

## 2.4 The existing negation method in probability theory and Dempster-Shafer evidence theory

The concept of "negation" has been applied in a number of theories, such as probability theory, complex-valued evidence theory, evidence theory, quantum evidence theory, discrete Z-number theory and so on. Since the permutation set theory is the generalization of probability theory and evidence theory, Yager's negation of probability distribution and Yin et al's negation of mass function are briefly introduced as follow.

### 2.3.1 Yager's negation of probability distribution

Supposing that a probability distribution is $P=\{p_1,p_2,\cdots p_n\}$, the corresponding negation of $p_i$ is defined as $\bar{p}_i = \dfrac{1-p_i}{n-1} = \dfrac{1-p_i}{\sum_{i=1}^{n}1-p_i}$. It can be find that $\sum_{i=1}^{n}\bar{p}_i=1$ and $n$-1 represents normalization factor.

### 2.3.2 Yin et al's negation of mass function

Supposing that a BPA defined on $\Phi=\{\varphi_1,\varphi_2,\cdots\varphi_N\}$ is

$\{m(\emptyset), m(A_1), m(A_2), \cdots m(A_{2^N-1})\}$, where $A \in 2^\Phi$, the corresponding negation of $m(A_i)$ is $\overline{m}(A_i) = \dfrac{1-m(A_i)}{n-1} = \dfrac{1-m(A_i)}{\sum_{i=1}^{n} 1-m(A_i)}$. It should be noted that $n$ is the number of focal element and there exists $\sum_{i=1}^{n} \overline{m}(A_i) = 1$.

### 2.5 The distance of random permutation sets

In order to measure the of dissimilarity degree between two *PM*s, Chen et al[31] proposed the distance of random permutation sets, which is defined as follows.

$$d(\mathbf{PM}_1, \mathbf{PM}_2) = \sqrt{\frac{1}{2}(\mathbf{PM}_1 - \mathbf{PM}_2)\underline{\underline{\mathbf{RD}}}(\mathbf{PM}_1 - \mathbf{PM}_2)^T}$$

where $\mathbf{PM}_i$ is a vector defined as $\mathbf{PM}_i = [PM_i(A_1), PM_i(A_2), \cdots PM_i(A_\Delta)]$, $i=1, 2$.

$\underline{\underline{\mathbf{RD}}}$ is a $\Delta \times \Delta$ matrix whose elements are

$$\underline{\underline{\mathbf{RD}}} = \begin{pmatrix} \frac{|A_1 \cap A_1|}{|A_1 \cup A_1|} \times OD(A_1, A_1) & \frac{|A_1 \cap A_2|}{|A_1 \cup A_2|} \times OD(A_1, A_2) & \cdots & \frac{|A_1 \cap A_\Delta|}{|A_1 \cup A_\Delta|} \times OD(A_1, A_\Delta) \\ \frac{|A_2 \cap A_1|}{|A_2 \cup A_1|} \times OD(A_2, A_1) & \frac{|A_2 \cap A_2|}{|A_2 \cup A_2|} \times OD(A_2, A_2) & \cdots & \frac{|A_2 \cap A_\Delta|}{|A_2 \cup A_\Delta|} \times OD(A_2, A_\Delta) \\ \vdots & \vdots & \ddots & \vdots \\ \frac{|A_\Delta \cap A_1|}{|A_\Delta \cup A_1|} \times OD(A_\Delta, A_1) & \frac{|A_\Delta \cap A_2|}{|A_\Delta \cup A_2|} \times OD(A_\Delta, A_2) & \cdots & \frac{|A_\Delta \cap A_\Delta|}{|A_\Delta \cup A_\Delta|} \times OD(A_\Delta, A_\Delta) \end{pmatrix}$$

$OD(A_i, A_j)$ is the ordered degree between $A_i$ and $A_j$, which is calculated as

$$OD(A_i, A_j) = \exp\left(-\frac{\sum_{\theta \in A_i \cap A_j} |\text{rank}_{A_i}(\theta) - \text{rank}_{A_j}(\theta)|}{|A_i \cup A_j|}\right)$$

where $\text{rank}_{A_i}(\theta)$ and $\text{rank}_{A_j}(\theta)$ are the order of element $\theta$ in $A_i$ and $A_j$, respectively. For example, if $A_1 = \{\gamma_2, \gamma_3, \gamma_1\}$, $A_2 = \{\gamma_1, \gamma_2\}$, the order of $\gamma_1$ in $A_1$ is 3, the order of $\gamma_1$ in $A_2$ is 1, i.e., $\text{rank}_{A_1}(\gamma_1) = 3$, $\text{rank}_{A_2}(\gamma_1) = 1$.

### 2.6 The entropy of random permutation sets

In order to measure the uncertainty for a random permutation sets, Deng[32]

proposed the entropy of random permutation sets, which is defined as

$$H_{RPS}(PM) = -\sum_{i=1}^{N}\sum_{j=1}^{P(N,i)} PM(A_{ij}) \log_2 \left( \frac{PM(A_{ij})}{F(i)-1} \right)$$

where $F(i) = \sum_{k=0}^{i} P(i,k) = \sum_{k=0}^{i} \frac{i!}{(i-k)!}$. $H_{RPS}$ shows good compatibility with existing theories. It will degenerate back to Deng entropy and Shannon's entropy under certain circumstances.

## 3 The proposed negation method of *PM*

"Negation" is a novel perspective of knowledge representation. There are already various ways of to obtain "negation" in probability and evidence theory. However, the negation in random permutation sets theory has not been explored. According to the previous analysis, "negation" can be viewed as a reassignment of belief. Based on this idea, in this section, two different negation methods have been proposed within the framework of random permutation theory.

### 3.1 The proposed negation of permutation mass function

Let $\overline{PM(A_i)}$ be the negation of the permutation mass function $PM(A_i)$, it can be calculated as

$$\overline{PM(A_i)} = \frac{1-PM(A_i)}{\Delta - 2}$$

where $\Delta = \sum_{i=0}^{N} P(N,i)$ is the cardinality of $PES(\Theta)$. The negated *PM* assigns the belief on each $A \in PES(\Phi), A \neq \varnothing$.

### 3.2 The analysis of proposed negation---from the view of belief reassignment

As previous mentioned, the operation of negation is a belief reassignment in a specific event space. When the negation operation is performed using the proposed method, the belief will be assigned to all events on the $PES(\Theta)$ except the empty set, regardless of whether an event is a focal element or not. The following example is used to illustrate the belief reassignment.

*Example 1* Assuming that $PES(\Theta) = \{\gamma_1, \gamma_2\}$, the permutation mass function defined on $\Phi$ is $PM(\gamma_1) = 0.1, PM(\gamma_2) = 0.7, PM(\gamma_1, \gamma_2) = 0.2, PM(\gamma_2, \gamma_1) = 0$. The negation of *PM* by using proposed method is calculated as:

$$\overline{PM(\gamma_1)} = \frac{1-PM(\gamma_1)}{4-1} = \frac{1-0.1}{3} = 0.3$$

$$\overline{PM(\gamma_2)} = \frac{1-PM(\gamma_2)}{4-1} = \frac{1-0.7}{3} = 0.1$$

$$\overline{PM(\gamma_1,\gamma_2)} = \frac{1-PM(\gamma_1,\gamma_2)}{4-1} = \frac{1-0.2}{3} = 0.2667$$

$$\overline{PM(\gamma_2,\gamma_1)} = \frac{1-PM(\gamma_2,\gamma_1)}{4-1} = \frac{1-0}{3} = 0.3333$$

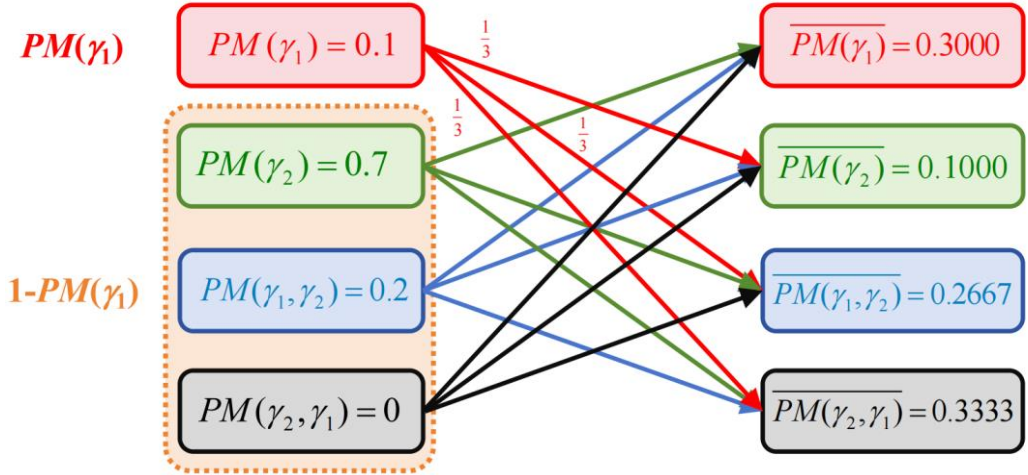

Figure 2 The belief reassignment process

Figure 2 visually shows the process of the belief reassignment. The analysis will be conducted using $PM(\gamma_1)$ and its negation as an example. Firstly, $PM(\gamma_1)$ is divided into three equal parts and has been allocated to $\overline{PM(\gamma_2)}$, $\overline{PM(\gamma_1,\gamma_2)}$, and $\overline{PM(\gamma_2,\gamma_1)}$ respectively. Secondly, the calculation of $\overline{PM(\gamma_1)}$ can be seen as $\overline{PM(\gamma_1)} = \frac{1}{3}PM(\gamma_2) + \frac{1}{3}PM(\gamma_1,\gamma_2) + \frac{1}{3}PM(\gamma_2,\gamma_1)$. Therefore, when proposed method is used to take the negation of original *PM*, for each $A \in PES(\Theta), A \neq \varnothing$ (whether *A* is a focal element or not), it will participate in this belief redistribution process.

Assuming that *PM* is regarded as a $(\Delta-1)$-dimension vector and do not consider the empty set $\varnothing$. According to example 1, when negation operation is applied, original *PM* assigns the components of any one dimension to all of the remaining dimensions.

# 4 Convergence, entropy and dissimilarity of *PM* during multiple negation operations

Consider the following negation iteration process: denote the initial permutation mass function as $PM_0$, the negation of $PM_0$ is denoted as $PM_1$, the negation of $PM_1$ is denoted as $PM_2$......after $i$th negation operations on $PM_0$, it is denoted as $PM_i$. Assuming that $\Theta = \{A, B\}$,

$PM_0 : PM_0(A) = 0.1, PM_0(B) = 0.7, PM_0(A,B) = 0.7, PM_0(B,A) = 0$. During this negation iteration process, the convergence, entropy and dissimilarity will be investigated. In this section, nine consecutive negation operations are performed on $PM_0$.

## 4.1 From the view of convergence

The value of $PM_i$ by using proposed method are presented in Table 1 and Figure 3. According to Figure 3, the following conclusions can be obtained: (1)when the number of negations is increasing, $PM_i(A)$, $PM_i(B)$, $PM_i(A, B)$ and $PM_i(B, A)$ will converge to a fixed value 0.2500. (2)The original $PM(PM_0)$ is not equal to the one after taking the negation operation twice($PM_2$). In other words, the negation process is "irreversible".

Table 1 The value of $PM_i$ by using proposed negation method

| $i$ | $PM_i(A)$ | $PM_i(B)$ | $PM_i(A, B)$ | $PM_i(B, A)$ |
|---|---|---|---|---|
| 0 | 0.1000 | 0.7000 | 0.2000 | 0 |
| 1 | 0.3000 | 0.1000 | 0.2667 | 0.3333 |
| 2 | 0.2333 | 0.3000 | 0.2444 | 0.2222 |
| 3 | 0.2556 | 0.2333 | 0.2519 | 0.2593 |
| 4 | 0.2481 | 0.2556 | 0.2494 | 0.2469 |
| 5 | 0.2506 | 0.2481 | 0.2502 | 0.2510 |
| 6 | 0.2498 | 0.2506 | 0.2499 | 0.2497 |
| 7 | 0.2501 | 0.2498 | 0.2500 | 0.2501 |
| 8 | 0.2500 | 0.2501 | 0.2500 | 0.2499 |
| 9 | 0.2500 | 0.2500 | 0.2500 | 0.2500 |

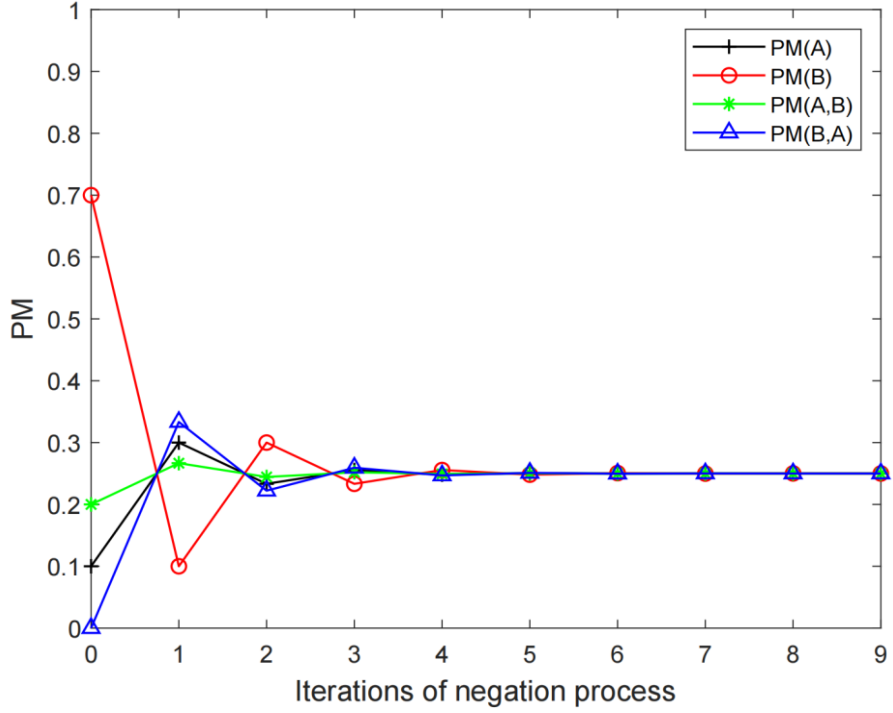

Figure 3 The change of $PM_i$ by using proposed method

Then the convergence of $PM_i$ in the negation process is further analyzed. For proposed negation method, there exists $PM_{i+1} = \frac{1 - PM_i}{\Delta - 2}$. When $i \to \infty$, $PM_i$ will converge to $\frac{1}{\Delta - 1}$. The proof is as follow.

Based on the negation method 1 and negation method 2, we have

$$PM_{i+1} = \frac{1 - PM_i}{\Delta - 2}.$$

Then

$$\begin{aligned} PM_{i+1} - \frac{1}{\Delta - 1} &= \frac{1 - PM_i}{\Delta - 2} - \frac{1}{\Delta - 1} \\ &= \frac{1}{\Delta - 2} - \frac{PM_i}{\Delta - 2} - \frac{1}{\Delta - 1} \\ &= \frac{1}{\Delta - 2} - \frac{(\Delta - 1)PM_i + (\Delta - 2)}{(\Delta - 2)(\Delta - 1)} \\ &= \left(-\frac{1}{\Delta - 2}\right)\left(PM_i - \frac{1}{\Delta - 1}\right) \end{aligned}$$

That is

$$\frac{PM_{i+1} - \frac{1}{\Delta-1}}{PM_i - \frac{1}{\Delta-1}} = -\frac{1}{\Delta-2}$$

Assume that

$$PM_i - \frac{1}{\Delta-1} = h_i$$

$$PM_{i+1} - \frac{1}{\Delta-1} = h_{i+1}$$

It can be concluded that $h_i$ is a geometric sequence with common ratio $-\frac{1}{\Delta-2}$.

Since $h_0 = PM_0 - \frac{1}{\Delta-1}$, The general term formula of the sequence $h_i$ is given by

$$h_i = \left(PM_0 - \frac{1}{\Delta-1}\right) \cdot \left(-\frac{1}{\Delta-2}\right)^i$$

Then

$$PM_i - \frac{1}{\Delta-1} = \left(PM_0 - \frac{1}{\Delta-1}\right) \cdot \left(-\frac{1}{\Delta-2}\right)^i$$

$$PM_i = \left(PM_0 - \frac{1}{\Delta-1}\right) \cdot \left(-\frac{1}{\Delta-2}\right)^i + \frac{1}{\Delta-1}$$

Since $\Delta - 2 > 1$, it is obvious that $\lim_{i \to \infty} PM_i = \frac{1}{\Delta-1}$.

It is obvious that the negation operation will tend to average the belief assignment.

**4.2 From the view of entropy**

According to the previous analysis, in this negation process, the operation of "negation" is irreversible. In other words, the value obtained after two consecutive negation operations on a *PM* is not equal to its initial value, which may be caused by the change of uncertainty during the negation process. In this section, Deng et al's entropy[32] based on random permutation sets theory $H_{RPS}$ is used to measure the uncertainty change in the negation process. The value of $H_{RPS}$ during the negation process by using different negation method is presented in Table 2 and Fig 4.

Table 2 The change of $H_{RPS}$ during the negation process

| i | $H_{RPS}$ |
|---|---|
| 0 | 1.5567 |
| 1 | 3.0901 |
| 2 | 2.9231 |
| 3 | 3.0212 |
| 4 | 2.9924 |
| 5 | 3.0023 |
| 6 | 2.9992 |
| 7 | 3.0002 |
| 8 | 3.0000 |
| 9 | 3.0000 |

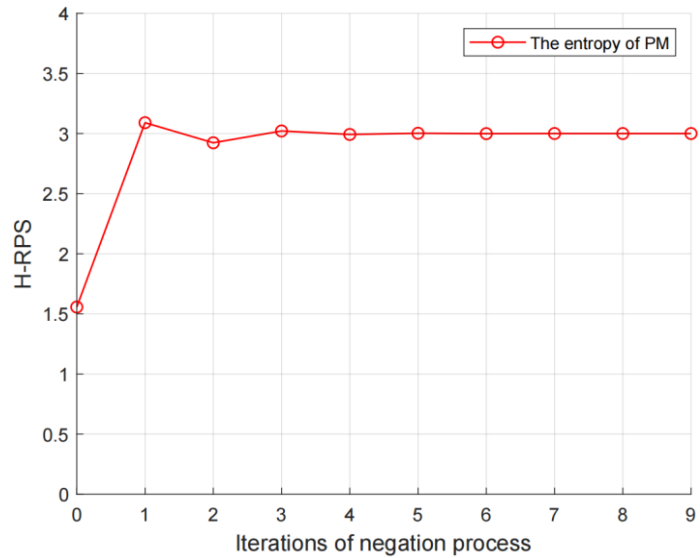

Fig.4 The change of $H_{RPS}$ during the negation process

From Fig.4, after the first negation operation, the $H_{RPS}$ will be significantly higher. In the subsequent negation process, the $H_{RPS}$ will fluctuate around a relatively high value and gradually converge to a fixed value. It should be highlighted that the convergence of $H_{RPS}$ results from the convergence of $PM_i$.

### 4.3 From the view of dissimilarity(distance)

According to the analysis in Section 4.1, in the process of negation, the value of $PM_i$ converges gradually, and its fluctuation becomes smaller and smaller. In other words, $PM_i$ and $PM_{i+1}$ are closer and closer to each other. This phenomenon can be verified by studying the variation trend of the distance between $PM_i$ and $PM_{i+1}$. In this section, Deng et al's distance measure[31] is used to represent the dissimilarity(conflict) of $PM_i$ and $PM_{i+1}$. The value of $d(PM_i, PM_{i+1})$ by using proposed negation method is shown in Table 3.

Table 3 The value of $d(PM_i, PM_{i+1})$ in the negation process

| $i$ | $d(PM_i, PM_{i+1})$ |
|---|---|
| 0 | 0.6633 |
| 1 | 0.2211 |
| 2 | 0.0737 |
| 3 | 0.0245 |
| 4 | 0.0081 |
| 5 | 0.0027 |
| 6 | 0.0009 |
| 7 | 0.0003 |
| 8 | 0.0001 |

It can be easily find that $d(PM_i, PM_{i+1})$ can be seen as a geometric sequence with common ratio $\frac{1}{3}$. Generally speaking, there exists $\frac{d_{RPS}(PM_{i+1}, PM_{i+2})}{d_{RPS}(PM_i, PM_{i+1})} = K$, $\lim_{i \to \infty} d_{RPS}(PM_i, PM_{i+1}) = 0$. The proof is as follows.

According to Section 2.5, the distance between $PM_1$ and $PM_2$ is defined as

$$d_{RPS}(\overrightarrow{PM_1}, \overrightarrow{PM_2}) = \sqrt{\frac{1}{2}(\overrightarrow{PM_1} - \overrightarrow{PM_2})^T \underline{\underline{RD}}(\overrightarrow{PM_1} - \overrightarrow{PM_2})}$$

It can be transformed as

$$d_{RPS}(PM_1, PM_2) = \sqrt{\frac{1}{2} \sum_{r=1}^{\Delta} \sum_{s=1}^{\Delta} (PM_1(A_r) - PM_2(A_r))(PM_1(A_s) - PM_2(A_s)) \frac{|A_r \cap A_s|}{|A_r \cup A_s|} \times OD(A_r, A_s)}$$

where $A_r, A_s \in PES(\Theta)$.

Since

$$PM_{i+1}(A_r) = \frac{1 - PM_i(A_r)}{\Delta - 2}$$

thus

$$PM_i(A_r) - PM_{i+1}(A_r) = \frac{(\Delta-1)PM_i(A_r) - 1}{\Delta - 2}$$

$$PM_i(A_s) - PM_{i+1}(A_s) = \frac{(\Delta-1)PM_i(A_s) - 1}{\Delta - 2}$$

then

$$d_{\text{RPS}}(PM_i, PM_{i+1}) = \sqrt{\frac{1}{2}\sum_{r=1}^{\Delta}\sum_{s=1}^{\Delta} \frac{[(\Delta-1)PM_i(A_r)-1][(\Delta-1)PM_i(A_s)-1]}{(\Delta-2)^2} \frac{|A_r \cap A_s|}{|A_r \cup A_s|} \times OD(A_r, A_s)}$$

Similarly, it can be obtained that

$$PM_{i+1}(A_r) - PM_{i+2}(A_r) = PM_{i+1}(A_r) - \frac{1 - PM_{i+1}(A_r)}{\Delta - 2}$$
$$= \frac{1 - PM_i(A_r)}{\Delta - 2} - \frac{PM_i(A_r) + \Delta - 3}{(\Delta - 2)^2}$$
$$= \frac{1 - (\Delta - 1)PM_i(A_r)}{(\Delta - 2)^2}$$

Then

$$d_{\text{RPS}}(PM_{i+1}, PM_{i+2}) = \sqrt{\frac{1}{2}\sum_{r=1}^{\Delta}\sum_{s=1}^{\Delta} \frac{[(\Delta-1)PM_i(A_r)-1][(\Delta-1)PM_i(A_s)-1]}{(\Delta-2)^4} \frac{|A_r \cap A_s|}{|A_r \cup A_s|} \times OD(A_r, A_s)}$$

Therefore, we have

$$\frac{d_{\text{RPS}}(PM_{i+1}, PM_{i+2})}{d_{\text{RPS}}(PM_i, PM_{i+1})} = \frac{1}{\Delta - 2}$$

Assuming that $d_{\text{RPS}}(PM_0, PM_1) = d_0$, there exists

$$d_{\text{RPS}}(PM_i, PM_{i+1}) = \frac{d_0}{(\Delta - 2)^i}$$

$$\lim_{i \to \infty} d_{\text{RPS}}(PM_i, PM_{i+1}) = 0$$

The above proof explains the decreasing trend of $d(PM_i, PM_{i+1})$, which is a new perspective to investigate the convergence of $PM_i$.

## Conclusion

Negation is a important way for knowledge representation. At present, the negation method in the framework of probability theory and evidence theory has been proposed, but the negation method in the framework of random permutation sets theory has not been studied. In this paper, two negation methods in random permutation sets theory is proposed. In this paper, the feasibility of the proposed two methods is analyzed in terms of belief redistribution, convergence, uncertainty (entropy), and dissimilarity (distance), and the similarities and differences of the two inversion methods are analyzed. How to design a more reasonable negation method within the framework of random permutation sets requires further investigation.